\documentclass[10pt,journal,compsoc]{IEEEtran}
\usepackage[american]{babel}

\usepackage{graphicx}
\usepackage{amsmath}
\usepackage{amssymb} 
\usepackage{mathrsfs}
\usepackage{multicol,multirow}

\usepackage{bm}
\usepackage{booktabs}
\usepackage[ruled,linesnumbered]{algorithm2e}
\usepackage{subcaption}
\usepackage{lipsum}
\usepackage{colortbl}
\usepackage{overpic}
\definecolor{mygray}{gray}{.8}

\usepackage[pagebackref=false,breaklinks=false,letterpaper=false,colorlinks,bookmarks=false]{hyperref}

\def\ie{\emph{i.e.}}
\def\eg{\emph{e.g.}}

\newcommand{\rev}[1]{\textcolor{red}{#1}}
\newcommand{\blu}[1]{\textcolor{blue}{#1}}

\usepackage{wrapfig}
\usepackage{tikz,pgfplots}

\usepackage{ragged2e}
\renewcommand{\raggedright}{\leftskip=0pt \rightskip=0pt plus 0cm}
  
\ifCLASSOPTIONcompsoc
  \usepackage[nocompress]{cite}
\else
  \usepackage{cite}
\fi

\ifCLASSINFOpdf
\else
\fi

\begin{document}
%


\title{Can SAM Segment Polyps?}

 \author{Tao Zhou, ~~Yizhe Zhang, ~~Yi Zhou, ~~Ye Wu, ~~Chen Gong
 	\\

\IEEEcompsocitemizethanks{
\IEEEcompsocthanksitem T. Zhou, Y. Zhang, Y. Wu, and C. Gong are with the School of Computer Science and Engineering, Nanjing University of Science and Technology, Nanjing, China. Y. Zhou is with the School of Computer Science and Engineering, Southeast University, Nanjing, China.
}

}

\markboth{IEEE Transactions on Pattern Analysis and Machine Intelligence}%
{Shell \MakeLowercase{\textit{et al.}}: Bare Demo of IEEEtran.cls for Computer Society Journals}


\markboth{}%
{Shell \MakeLowercase{\textit{et al.}}: Bare Demo of IEEEtran.cls for Computer Society Journals}

\IEEEtitleabstractindextext{
\begin{abstract}
\raggedright{
Recently, Meta AI Research releases a general Segment Anything Model (SAM), which has demonstrated promising performance in several segmentation tasks. As we know, polyp segmentation is a fundamental task in the medical imaging field, which plays a critical role in the diagnosis and cure of colorectal cancer. In particular, applying SAM to the polyp segmentation task is interesting. In this report, we evaluate the performance of SAM in segmenting polyps, in which SAM is under unprompted settings. We hope this report will provide insights to advance this polyp segmentation field and promote more interesting works in the future. This project is publicly at \href{https://github.com/taozh2017/SAMPolyp}{https://github.com/taozh2017/SAMPolyp}.
}

\end{abstract}

}

\maketitle




\section{Introduction}

\IEEEPARstart{l}{arge} Language Models (LLMs), such as ChatGPT~\cite{brown2020language} and GPT-4~\cite{chat4}, are pre-trained on vast amounts of text data and have demonstrated impressive performance in various natural language processing (NLP) tasks. The effectiveness of LLMs leads people to consider that how to effectively utilize these models in real-world applications. However, they are not specifically designed for image segmentation tasks \cite{kirillov2023segment}.



Recently, Segment Anything Model (SAM)~\cite{kirillov2023segment} is released as a general image segmentation model, which is trained on the large visual corpus (SA-1B). SAM produces high-quality object masks from input prompts (\eg, points, boxes, and masks). Due to its promising performance in several segmentation tasks, SAM has attracted increasing attention to applying it to various fields. For example, Ji \emph{et al.}~\cite{jiwei2023} discussed the benefits and limitations of SAM in different computer vision and medical image segmentation tasks. Tang \emph{et al.}~\cite{tang2023can} evaluated the performance of SAM in the field of Camouflaged Object Detection (COD). Ji~\emph{et al.}~\cite{ji2023} evaluated SAM on three concealed scenes (\ie, camouflaged animals, industrial defects, and medical lesions) under unprompted settings. Roy~\emph{et al.}~\cite{roy2023sam} valuated the zero-shot capabilities of SAM on the organ segmentation task. Deng~\emph{et al.}~\cite{deng2023segment} applied SAM to segment heterogeneous objects in digital pathology. The above studies show SAM can not perform well under several real-world applications. Among medical image segmentation tasks, polyp segmentation is a critical step in the diagnosis and cure of colorectal cancer. Thus, it is of great interest to investigate how well SAM can segment polyps in colonoscopy images under different challenging factors.

This report mainly evaluates the effectiveness of SAM in segmenting polyps. To achieve this, we compare SAM quantitatively and qualitatively with state-of-the-art models on the polyp segmentation task. Under the unprompted setting, it can be seen that SAM does not perform better in the polyp segmentation task. Therefore, how to effectively apply SAM to the polyp segmentation should be further explored.

\begin{table*}[tp]
  \centering
  \small
  \renewcommand{\arraystretch}{1.2}
  \renewcommand{\tabcolsep}{1.8mm}
  \caption{Quantitative polyp segmentation results on the CVC-ClinicDB and Kvasir datasets. $\dag$ denotes a Transformer-based model.} \vspace{-0.2cm}
  \begin{tabular}{r||cccccc|cccccc}
  \hline
    \multirow{2}*{\textbf{Methods}}   
    &\multicolumn{6}{c|}{CVC-ClinicDB \cite{bernal2015wm}}
    &\multicolumn{6}{c}{Kvasir~\cite{jha2020kvasir}}\\

     \cline{2-13}
  
    & mDice $\uparrow$  & mIou $\uparrow$  & $S_{\alpha}\uparrow$  &$F_{\beta}^{w}\uparrow$ & $E_{\phi}^{max}\uparrow$  &$\mathcal{M}\downarrow$
    & mDice $\uparrow$  & mIou $\uparrow$  & $S_{\alpha}\uparrow$  &$F_{\beta}^{w}\uparrow$ & $E_{\phi}^{max}\uparrow$  &$\mathcal{M}\downarrow$ \\

    \hline
  
  
    UNet~\cite{ronneberger2015u}~
    & 0.823   & 0.755   & 0.889 & 0.811   & 0.954 & 0.019	
    & 0.818   & 0.746   & 0.858 & 0.794   & 0.893 & 0.055 \\
    
    UNet++~\cite{zhou2019unet++}~
    & 0.794   & 0.729   & 0.873 & 0.785   & 0.931 & 0.022	
    & 0.821   & 0.744   & 0.862 & 0.808   & 0.910 & 0.048 \\

    SFA~\cite{fang2019selective}~
    & 0.700   & 0.607   & 0.793 & 0.647   & 0.885 & 0.042	
    & 0.723   & 0.611   & 0.782 & 0.670   & 0.849 & 0.075\\   

    PraNet~\cite{fan2020pranet}~
    & 0.899   & 0.849   & 0.936 & 0.896   & 0.979  & 0.009	
    & 0.898   & 0.840   & 0.915 & 0.885   & 0.948  & 0.030\\  

    ACSNet~\cite{zhang2020adaptive}~
    & 0.882   & 0.826   & 0.927 & 0.873   & 0.959 & 0.011	
    & 0.898   & 0.838   & 0.920 & 0.882   & 0.952 & 0.032 \\
    
    MSEG\cite{huang2021hardnet}~
    & 0.909   & 0.864   & 0.938 & 0.907   & 0.969 & 0.007	
    & 0.897   & 0.839   & 0.912 & 0.885   & 0.948 & 0.028 \\

    DCRNet~\cite{yin2022duplex}~
    & 0.896   & 0.844   & 0.933 & 0.890   & 0.978 & 0.010	
    & 0.886   & 0.825   & 0.911 & 0.868   & 0.941 & 0.035\\   

    EU-Net~\cite{patel2021enhanced}~
    & 0.902   & 0.846   & 0.936 & 0.891   & 0.965  & 0.011	
    & 0.908   & 0.854   & 0.917 & 0.893   & 0.954  & 0.028\\  

    SANet~\cite{wei2021shallow}~
    & 0.916   & 0.859   & 0.939 & 0.909   & 0.976  & 0.012	
    & 0.904   & 0.847   & 0.915 & 0.892   & 0.953  & 0.028\\  

    MSNet~\cite{zhao2021automatic}~
    & 0.918   & 0.869   & 0.946  & 0.913   & 0.979  & 0.008	
    & 0.905   & 0.849   & 0.923  & 0.892   & 0.954  & 0.028\\  
     
    C2FNet~\cite{sun2021context}~
    & 0.919   & 0.872  & 0.941 & 0.906   & 0.976  & 0.009	
    & 0.886   & 0.831  & 0.905 & 0.870   & 0.935  & 0.036\\  


    LDNet~\cite{zhang2022lesion}~
    & 0.881   & 0.825  & 0.924 & 0.879   & 0.965  & 0.012	
    & 0.887   & 0.821  & 0.905 & 0.869   & 0.945  & 0.031\\ 

    FAPNet~\cite{zhou2022feature}~
    & 0.925   & 0.877  & 0.947  & 0.910   & 0.979  & 0.008
    & 0.902   & 0.849  & 0.919  & 0.894   & 0.953  & \blu{0.027}\\

    CFA-Net~\cite{zhou2023cross}~
    & 0.933   & 0.883   & 0.950 & 0.924   & \blu{0.989}  & \blu{0.007}	
    & 0.915   & 0.861   & 0.924 & 0.903   & \blu{0.962}  & \rev{0.023}\\

    \hline
    
    Polyp-PVT$^{\dag}$~\cite{dong2021polyp}~
    & \rev{0.948}   & \rev{0.905}   & \rev{0.953}  & \rev{0.951}   & \rev{0.993}  & \rev{0.006}	
    & \blu{0.917}   & \blu{0.864}   & \blu{0.925}  & \blu{0.911}   & \blu{0.962}  & \rev{0.023}\\

    HSNet$^{\dag}$~\cite{zhang2022hsnet}~
    & \blu{0.937}   & \blu{0.887}   & \blu{0.949} & \blu{0.936}   & \blu{0.989}  & \rev{0.006}	
    & \rev{0.926}   & \rev{0.877}   & \rev{0.927} & \rev{0.918}   & \rev{0.964}  & \rev{0.023}\\
    
    \hline \hline
    SAM-H~\cite{kirillov2023segment}~
    & 0.547   & 0.500   & 0.738  & 0.546   & 0.677  & 0.040	
    & 0.778   & 0.707   & 0.829  & 0.769   & 0.831  & 0.062\\

    SAM-L~\cite{kirillov2023segment}~
    & 0.579   & 0.526   & 0.744  & 0.563   & 0.685  & 0.057	
    & 0.782   & 0.710   & 0.832  & 0.773   & 0.836  & 0.061\\


  \hline
  \end{tabular}\label{tab1}
\end{table*}

\begin{table*}[tp]
  \centering
  \small
  \renewcommand{\arraystretch}{1.2}
  \renewcommand{\tabcolsep}{1.8mm}
  \caption{Quantitative polyp segmentation results on the CVC-300 and ColonDB datasets. $\dag$ denotes a Transformer-based model.} \vspace{-0.2cm}
  \begin{tabular}{r||cccccc|cccccc}
  \hline
    \multirow{2}*{\textbf{Methods}}   
    &\multicolumn{6}{c|}{CVC-300 \cite{vazquez2017benchmark}}
    &\multicolumn{6}{c}{ColonDB~\cite{tajbakhsh2015automated}}\\

     \cline{2-13}
  
    & mDice $\uparrow$  & mIou $\uparrow$  & $S_{\alpha}\uparrow$  &$F_{\beta}^{w}\uparrow$ & $E_{\phi}^{max}\uparrow$  &$\mathcal{M}\downarrow$
    & mDice $\uparrow$  & mIou $\uparrow$  & $S_{\alpha}\uparrow$  &$F_{\beta}^{w}\uparrow$ & $E_{\phi}^{max}\uparrow$  &$\mathcal{M}\downarrow$ \\

    \hline 
  
  
    UNet~\cite{ronneberger2015u}~
    & 0.710   & 0.627   & 0.843 & 0.684   & 0.876 & 0.022	
    & 0.504   & 0.436   & 0.710 & 0.491   & 0.781 & 0.059 \\
    
    UNet++~\cite{zhou2019unet++}~
    & 0.707   & 0.624   & 0.839 & 0.687   & 0.898 & 0.018	
    & 0.482   & 0.408   & 0.693 & 0.467   & 0.764 & 0.061 \\

    SFA~\cite{fang2019selective}~
    & 0.467   & 0.329   & 0.640 & 0.341   & 0.817 & 0.065	
    & 0.456   & 0.337   & 0.629 & 0.366   & 0.754 & 0.094\\   

    PraNet~\cite{fan2020pranet}~
    & 0.871   & 0.797   & 0.925 & 0.843   & 0.972  & 0.010	
    & 0.712   & 0.640   & 0.820 & 0.699   & 0.872  & 0.043\\  

    ACSNet~\cite{zhang2020adaptive}~
    & 0.863   & 0.787   & 0.923 & 0.825   & 0.968 & 0.013	
    & 0.716   & 0.649   & 0.829 & 0.697   & 0.851 & 0.039 \\
    
    MSEG\cite{huang2021hardnet}~
    & 0.874   & 0.804   & 0.924  & 0.852   & 0.957 & 0.009	
    & 0.735   & 0.666   & 0.834  & 0.724   & 0.875 & 0.038 \\

    DCRNet~\cite{yin2022duplex}~
    & 0.856   & 0.788   & 0.921  & 0.830   & 0.960 & 0.010	
    & 0.704   & 0.631   & 0.821  & 0.684   & 0.848 & 0.052\\   

    EU-Net~\cite{patel2021enhanced}~
    & 0.837   & 0.765   & 0.904  & 0.805   & 0.933  & 0.015	
    & 0.756   & 0.681   & 0.831  & 0.730   & 0.872  & 0.045\\      
    
    SANet~\cite{wei2021shallow}~
    & 0.888   & 0.815   & 0.928 & 0.859   & 0.972  & \blu{0.008}	
    & 0.753   & 0.670   & 0.837 & 0.726   & 0.878  & 0.043\\  

    MSNet~\cite{zhao2021automatic}~
    & 0.865   & 0.799   & 0.926  & 0.848   & 0.953  & 0.010	
    & 0.751   & 0.671   & 0.838  & 0.736   & 0.883  & 0.041\\ 
    
    C2FNet~\cite{sun2021context}~
    & 0.874   & 0.801    & 0.927 & 0.844   & 0.968  & 0.009
    & 0.724   & 0.650   & 0.826 & 0.705   & 0.868  & 0.044\\  
        

    LDNet~\cite{zhang2022lesion}~
    & 0.869   & 0.793  & 0.923 & 0.841   & 0.965  & 0.009	
    & 0.740   & 0.652  & 0.830 & 0.717   & 0.884  & 0.036\\  

    FAPNet~\cite{zhou2022feature}~
    & 0.893   & 0.826  & \rev{0.938}   & 0.874  & 0.976  & \blu{0.008}
    & 0.731   & 0.658  & 0.831   & 0.735  & 0.878 & 0.038\\

    CFA-Net~\cite{zhou2023cross}~
    & 0.893   & 0.827   & \rev{0.938} & 0.875   & \blu{0.978}  & \blu{0.008}	
    & 0.743   & 0.665   & 0.835 & 0.728   & 0.898  & 0.039	\\

    \hline
    Polyp-PVT$^{\dag}$~\cite{dong2021polyp}~
    & \blu{0.900}   & \blu{0.833}   & 0.935 & \blu{0.884}   & \rev{0.981}  & \rev{0.007}	
    & \blu{0.808}   & \blu{0.727}   & 0.\blu{865} & \blu{0.795}   & \rev{0.919}  & \rev{0.031}\\

    HSNet$^{\dag}$~\cite{zhang2022hsnet}~
    & \rev{0.903}   & \rev{0.839}   & \blu{0.937} & \rev{0.887}   & 0.975  & \rev{0.007}	
    & \rev{0.810}   & \rev{0.735}   & \rev{0.868} & \rev{0.796}   & \blu{0.915}  & \blu{0.032}	\\
    
    \hline
    \hline
    SAM-H~\cite{kirillov2023segment}~
    & 0.651   & 0.606   & 0.812 & 0.653   & 0.767  & 0.020	
    & 0.441   & 0.396   & 0.676 & 0.434   & 0.587  & 0.056\\

    SAM-L~\cite{kirillov2023segment}~
    & 0.726   & 0.676   & 0.849 & 0.729   & 0.826  & 0.020	
    & 0.468   & 0.422   & 0.690 & 0.463   & 0.608  & 0.054\\


  \hline
  \end{tabular}\label{tab2}
\end{table*}

\begin{table*}[t]
  \centering
  \small
  \renewcommand{\arraystretch}{1.2}
  \renewcommand{\tabcolsep}{3.8mm}
  \caption{Results on the ETIS dataset. $\dag$ denotes a Transformer-based model.} \vspace{-0.2cm}
  \begin{tabular}{r||cccccc}
   \toprule

    & mDice $\uparrow$  & mIou $\uparrow$  & $S_{\alpha}\uparrow$  &$F_{\beta}^{w}\uparrow$ & $E_{\phi}^{max}\uparrow$  &$\mathcal{M}\downarrow$\\

    \midrule
  
    UNet~\cite{ronneberger2015u}~
    & 0.398   & 0.335   & 0.684 & 0.366   & 0.740 & 0.036	\\
    
    UNet++~\cite{zhou2019unet++}~
    & 0.401   & 0.344  & 0.683 & 0.390   & 0.776 & 0.035	\\

    SFA~\cite{fang2019selective}~
    & 0.297   & 0.217   & 0.557 & 0.231   & 0.633 & 0.109	\\   

    PraNet~\cite{fan2020pranet}~
    & 0.628   & 0.567   & 0.794 & 0.600   & 0.841  & 0.031	\\  

    ACSNet~\cite{zhang2020adaptive}~	
    & 0.578   & 0.509   & 0.754 & 0.530   & 0.764 & 0.059 \\
    
    MSEG~\cite{huang2021hardnet}~
    & 0.700   & 0.630   & 0.828 & 0.671   & 0.890 & 0.015 \\

    DCRNet~\cite{yin2022duplex}~
    & 0.556   & 0.496   & 0.736 & 0.506   & 0.773 & 0.096\\   

    EU-Net~\cite{patel2021enhanced}~	
    & 0.687   & 0.609   & 0.793 & 0.636   & 0.841  & 0.067\\        
    
    SANet~\cite{wei2021shallow}~
    & 0.750   & 0.654   & 0.849  & 0.685   & 0.897  & 0.015	\\  

    MSNet~\cite{zhao2021automatic}~
    & 0.723   & 0.652   & 0.845 & 0.677   & 0.890  & 0.020	\\  
    
    C2FNet~\cite{sun2021context}~
    & 0.699   & 0.624   & 0.827 & 0.668   & 0.875  & 0.022	\\  

    
    LDNet~\cite{zhang2022lesion}~
    & 0.645   & 0.551  & 0.788 & 0.600   & 0.847  & 0.023\\  
    FAPNet~\cite{zhou2022feature}~
    & 0.717   & 0.643   & 0.841   & 0.657   & 0.884 & 0.019 \\

    CFA-Net~\cite{zhou2023cross}~
    & 0.732   & 0.655  & 0.845 & 0.693   & 0.892  & \blu{0.014}	\\

    \hline
    Polyp-PVT$^{\dag}$~\cite{dong2021polyp}~
    & \blu{0.787}   & \blu{0.706}   & \blu{0.871} & \blu{0.750}   & \rev{0.910}  & \rev{0.013}\\	

    HSNet$^{\dag}$~\cite{zhang2022hsnet}~
    & \rev{0.808}   & \rev{0.734}   & \rev{0.882} & \rev{0.777}   & \blu{0.909}  & 0.021\\
    
    \midrule
    \hline
    SAM-H~\cite{kirillov2023segment}~
    & 0.517   & 0.477   & 0.730 & 0.513   & 0.660  & 0.029\\

    SAM-L~\cite{kirillov2023segment}~
    & 0.551   & 0.507   & 0.751 & 0.544   & 0.687  & 0.030\\


  \bottomrule
  \end{tabular}\label{tab3}
\end{table*}

\section{Experiments and Results}
\label{Experiments}

\subsection{Datasets}
\label{datasets}


To validate the effectiveness of different polyp segmentation models, we conduct comparison experiments on five benchmark colonoscopy datasets, and the details of each dataset are introduced below. $\bullet$~\textbf{Kvasir}~\cite{jha2020kvasir}: This dataset is collected by Vestre Viken Health Trust in Norway from inside the gastrointestinal tract, which
consists of $1,000$ polyp images. $\bullet$~\textbf{CVC-ClinicDB} \cite{bernal2015wm}: This dataset contains $612$ images collected from $29$ colonoscopy video sequences with a resolution of $288{\times}384$. $\bullet$~\textbf{CVC-ColonDB}~\cite{tajbakhsh2015automated}: This dataset consists of $380$ images with a resolution of $500\times{570}$. $\bullet$~\textbf{ETIS}~\cite{silva2014toward}: This dataset contains $196$ polyp images with a size of $966\times{1225}$. $\bullet$~\textbf{CVC-300}~\cite{vazquez2017benchmark}: This dataset includes $60$ polyp images with a resolution of $500{\times}574$. Following the same data split settings in \cite{fan2020pranet}, $900$ images from Kvasir and $550$ ones from CVC-ClinicDB are selected to form the training set. The remaining images from the two datasets (\ie, Kvasir and CVC-ClinicDB) and the other three datasets (\ie, CVC-ColonDB, ETIS, and CVC-300) are used for testing.

\begin{figure*}
\centering
\begin{overpic}[width=1.0\linewidth]{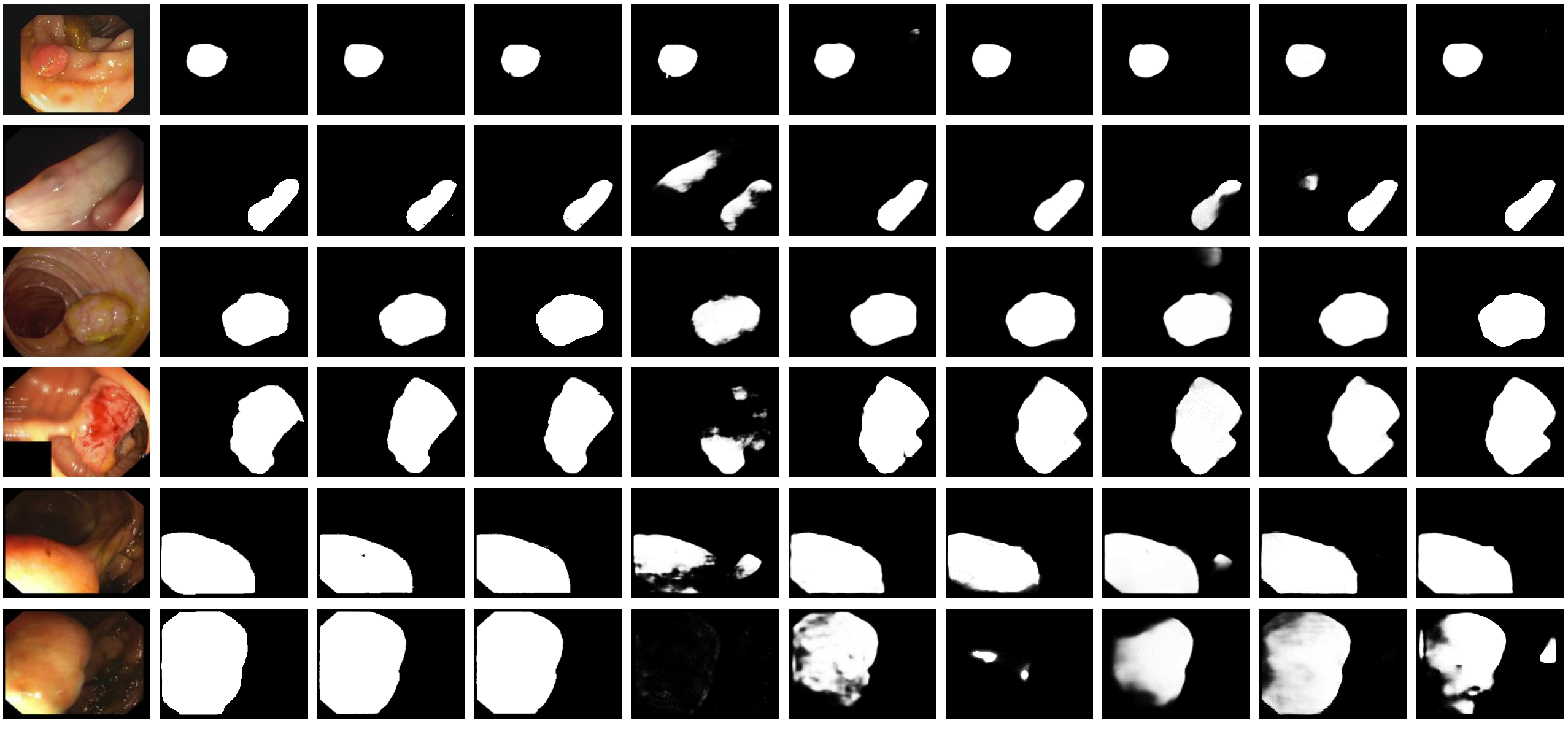}
\put(3, 0.1){\footnotesize Image}  
\put(13,0.1){\footnotesize GT}  
\put(22,0.1){\footnotesize SAM-H}  
\put(32,0.1){\footnotesize SAM-L}  
\put(42,0.1){\footnotesize UNet++}  
\put(52,0.1){\footnotesize MSEG}  
\put(62,0.1){\footnotesize SANet}  
\put(72,0.1){\footnotesize MSNet}  
\put(82,0.1){\footnotesize CFA-Net}
\put(91,0.1){\footnotesize Polyp-PVT}  
\end{overpic}
\caption{Some better segmentation examples of SAM.}  
 \label{fig01}
\end{figure*}

\subsection{Experimental Settings}
\label{setting}

\subsubsection{Comparison Methods}
In this study, we first evaluate 14 CNN-based polyp segmentation methods, including UNet~\cite{zhou2019unet++}, UNet++~\cite{zhou2019unet++}, SFA~\cite{fang2019selective}, PraNet~\cite{fang2019selective}, ACSNet~\cite{zhang2020adaptive}, MSEG\cite{huang2021hardnet}, DCRNet~\cite{yin2022duplex}, EU-Net~\cite{patel2021enhanced}, SANet~\cite{wei2021shallow}, MSNet~\cite{zhao2021automatic}, C2FNet~\cite{sun2021context}, LDNet~\cite{zhang2022lesion}, FAPNet~\cite{zhou2022feature}, and CFA-Net~\cite{zhou2023cross}. Besides, we evaluate two Transformer-based segmentation models, \ie, Polyp-PVT~\cite{dong2021polyp} and HSNet~\cite{zhang2022hsnet}. For SAM, we adopt two backbones~\cite{dosovitskiy2020image} in this study, denoted as ``SAM-H" and ``SAM-L", respectively.


\subsubsection{Evaluation Metrics and Setting} To evaluate the effectiveness of different models, we adopt six commonly used metrics \cite{yang2021mutual,zhou2021rgb,zhou2021specificity,fan2020camouflaged}, namely mean dice score (mDice), mean intersection over union (mIoU), S-measure ($S_{\alpha}$) \cite{fan2017structure}, F-measure \cite{achanta2009frequency} ($F_{\beta}^w$), $E_{\phi}^{max}$ \cite{Fan2018Enhanced}, mean absolute error ($\mathcal{M}$) \cite{perazzi2012saliency}. Due to the unprompted setting, SAM can produce multiple binary masks. Following these works~\cite{tang2023can,ji2023}, we adopt a strategy to select the best mask based on its ground-truth map. Specifically, we have obtained $N$ masks generated by SAM, denoted as $S_{i=1}^N$. For a given image, its ground-truth map is denoted as $S$. Thus, we compute $S_{\alpha}$ values for the $N$ masks, and the mask with the highest $S_{\alpha}$ score is selected as the segmentation map.

\begin{figure*}[!t]
\centering
\begin{overpic}[width=1.0\linewidth]{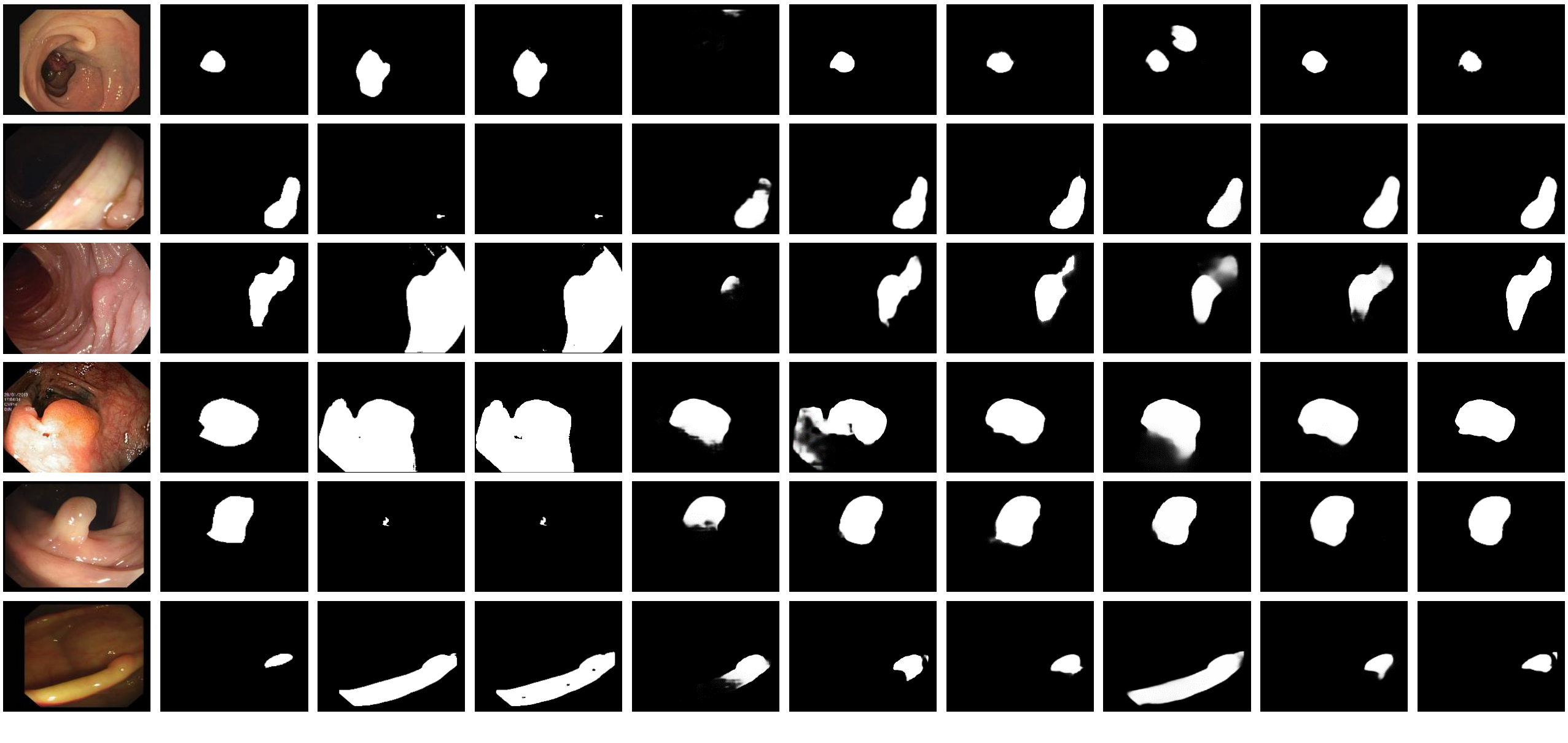}
\put(3, 0.1){\footnotesize Image}  
\put(13,0.1){\footnotesize GT}  
\put(22,0.1){\footnotesize SAM-H}  
\put(32,0.1){\footnotesize SAM-L}  
\put(42,0.1){\footnotesize UNet++}  
\put(52,0.1){\footnotesize MSEG}  
\put(62,0.1){\footnotesize SANet}  
\put(72,0.1){\footnotesize MSNet}  
\put(82,0.1){\footnotesize CFA-Net}
\put(91,0.1){\footnotesize Polyp-PVT}  
\end{overpic}
\caption{Some failure segmentation examples of SAM.}  
 \label{fig02}
\end{figure*}

\subsection{Segmentation Results}
\label{polyp}

\textbf{Quantitative Comparison:} Table~\ref{tab1}, Table~\ref{tab2}, and Table~\ref{tab3} show the quantitative results of different segmentation methods in the terms of six metrics on five datasets. From the results, we can observe that SAM can not effectively locate and segment polyps. On the one hand, polyp segmentation is challenging due to blurred boundaries between a polyp and its surrounding mucosa. On the other hand, SAM could be trained in a large number of natural images rather than medical images, which makes it challenging for SAM to segment polyps. 

\textbf{Qualitative Comparison:} We also show some better and bad segmentation examples of SAM in Fig.~\ref{fig01} and Fig.~\ref{fig02}, respectively. As shown in Fig.~\ref{fig01}, it can be seen that SAM accurately segments polyps in some scenes. However, SAM fails to segment polyps when the boundary between a polyp and its surrounding mucosa is non-sharp, as shown in the 3$^{nd}$ and 4$^{th}$ rows in Fig.~\ref{fig02}.

\section{Conclusion}
\label{conclusion}

This report provides a preliminary evaluation of SAM in segmenting polyps using colonoscopy images. We have conducted experiments on five benchmark datasets, and the results demonstrate there is still room for improvement when applying SAM to the polyp segmentation task. It can be noted that we directly apply the trained model from SAM to infer the polyp segmentation so that it is difficult to achieve satisfactory performance for these unseen medical images. Thus, one possible solution is to fine-tune the SAM model using training datasets of the specific task. In this case, we could obtain better segmentation performance than that without a fine-tuning strategy. We hope this report will promote more interest in this field of polyp segmentation, and more research works leveraging SAM will be developed in this area. To promote future research for polyp segmentation, we will collect awesome polyp segmentation models at: \href{https://github.com/taozh2017/Awesome-Polyp-Segmentation}{https://github.com/taozh2017/Awesome-Polyp-Segmentation}.

\bibliographystyle{IEEEtran}
\bibliography{sam_seg}

\end{document}